\documentclass[10pt,twocolumn,letterpaper]{article}

\usepackage[pagenumbers]{cvpr} 

\usepackage{graphicx}
\usepackage{amsmath}

\usepackage{mathtools}

\usepackage{graphicx}
\usepackage{multirow}
\usepackage{multicol}
\usepackage{colortbl}  

\usepackage{amssymb}
\usepackage{booktabs}
\usepackage{bigstrut}
\usepackage{multirow}
\usepackage{array}
\usepackage{amsthm}
\usepackage{diagbox}

\definecolor{cvprblue}{rgb}{0.21,0.49,0.74}
\definecolor{Gray}{gray}{0.86}

\newcommand{\rblue}{\rowcolor{blue!10}}

\definecolor{purple}{RGB}{112,48,160}
\definecolor{blue1}{RGB}{47,85,151}
\definecolor{orange}{RGB}{197,90,17}

\usepackage{marvosym}
\makeatletter
\def\@fnsymbol#1{\ensuremath{\ifcase#1\or \dagger\or \ddagger\or
   \mathsection\or \mathparagraph\or \|\or **\or \dagger\dagger
   \or \ddagger\ddagger \else\@ctrerr\fi}}
\makeatother

\usepackage[pagebackref,breaklinks,colorlinks,citecolor=cvprblue]{hyperref}

\def\paperID{4195} 
\def\confName{CVPR}
\def\confYear{2025}

\begin{document}

\title{Mixture of Physical Priors Adapter for Parameter-Efficient Fine-Tuning}

\author{
Zhaozhi Wang\textsuperscript{$1$}, Conghu Li\textsuperscript{$1$}, Qixiang Ye\textsuperscript{$1$}, Tong Zhang\textsuperscript{$1,2$}\thanks{Correspondence to \Letter \ tozhang.ucas@gmail.}
\\
\\
\textsuperscript{$1$}University of Chinese Academy of Sciences \quad \textsuperscript{$2$}EPFL}

\maketitle

\begin{abstract}

%

Most parameter-efficient fine-tuning (PEFT) methods rely on low-rank representations to adapt models. However, these approaches often oversimplify representations, particularly when the underlying data has high-rank or high-frequency components. This limitation hinders the model’s ability to capture complex data interactions effectively. In this paper, we propose a novel approach that models network weights by leveraging a combination of physical priors, enabling more accurate approximations. We use three foundational equations—heat diffusion, wave propagation, and Poisson’s steady-state equation—each contributing distinctive modeling properties: heat diffusion enforces local smoothness, wave propagation facilitates long-range interactions, and Poisson’s equation captures global equilibrium. To combine these priors effectively, we introduce the Mixture of Physical Priors Adapter (MoPPA), using an efficient Discrete Cosine Transform (DCT) implementation. To dynamically balance these priors, a route regularization mechanism is designed to adaptively tune their contributions. MoPPA serves as a lightweight, plug-and-play module that seamlessly integrates into transformer architectures, with adaptable complexity depending on the local context. 
Specifically, using MAE pre-trained ViT-B, MoPPA improves PEFT accuracy by up to 2.1\% on VTAB-1K image classification with a comparable number of trainable parameters, and advantages are further validated through experiments across various vision backbones, showcasing MoPPA's effectiveness and adaptability. The code will be made public available.

\end{abstract}

\section{Introduction}

With the growth in the size of modern models and the evolution of their pre-training techniques~\citep{gpt4,clip,mae,beit3,vit22b}, fine-tuning methods have recently undergone a notable paradigm shift.
Parameter-efficient fine-tuning (PEFT) has emerged as a key technique, outperforming conventional fine-tuning approaches when adapting large pre-trained models to target domains with limited training data~\citep{peftsurvey}.

Most existing PEFT methods retain only a minimal set of trainable weights while freezing the majority of a model’s parameters. These added weights are often structured with low-rank priors to limit complexity and parameter count~\citep{lora,sptlora,lorand}. While effective, they often require predefined rank sizes, constraining adaptability since different tasks and model layers may require distinct dimensionalities. Additionally, low-rank manifold priors might struggle with diverse pre-training datasets, especially where data distributions significantly differ between pre-training and fine-tuning stages. This calls for an adaptive approach that ensures flexible feature alignment without increasing parameters. Moreover, the inherent limitations of low-rank structures can restrict model capacity to capture complex, nuanced interactions in vision tasks, making them less effective for tasks requiring higher adaptability and precision.


\begin{figure}[t]
    \centering
    \includegraphics[width=0.98\linewidth]{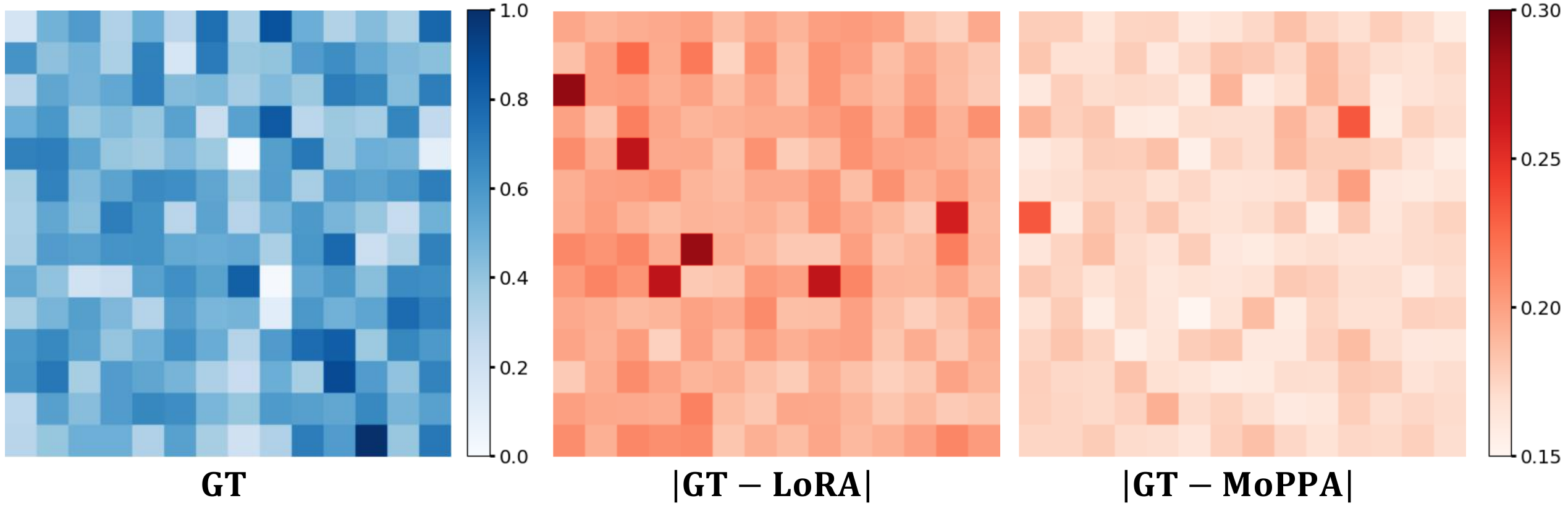}
    \caption{Visualization of the randomly generated Ground Truth (GT) and the absolute error between GT and regression results from LoRA / MoPPA. The results are averaged channel-wise. Please refer to Sec.~\ref{sec:analysis_implementation} in the supplementary for details on the adaptation analysis implementation.}
    \label{fig:moppa_error}
\end{figure}

In this paper, we address these challenges by replacing low-rank priors with physics-informed structures to construct trainable parameters, Fig.~\ref{fig:show}. Intuitively, we consider three different physical equations that can effectively approximate feature representation: the heat conduction equation~\citep{heateq} for localized feature adjustment, the wave propagation equation~\citep{waveeq} to extend receptive fields, and Poisson’s equation~\citep{possioneq} to capture potential fields influenced by electric charge distributions. As evidenced by Fig.~\ref{fig:moppa_error},  a linear combination of these priors provides a more accurate approximation than low-rank methods alone.

Hence, we propose a lightweight adapter, the Mixture of Physical Priors Adapter (MoPPA),which efficiently adapts image features for pre-trained models by leveraging a mixture of different physical equations. By grounding our approach in physical equations, we introduce an adaptable structure that can naturally vary in complexity depending on the local context within the model. Besides, the Physics-informed modeling enables feature transformations that are inherently robust to scale and structure. To simulate these physical transformations, we derive general solutions for them within the 2D space using discrete cosine transforms (DCT/IDCT). Given that the transformed features reside in the frequency domain, we assign learnable coefficients to each operator ($e.g.$, thermal diffusivity for the heat equation, wave speed for the wave equation, and density distribution for Poisson’s equation) based on frequency values.

To prevent premature convergence of the router’s path weights—a common risk which can lead to suboptimal solutions—we introduce a route regularization term in the training loss. This term discourages any early dominance of specific path weights and is gradually removed in later training stages to allow for stable optimization. Furthermore, we insert MoPPA units before pre-trained self-attention modules, promoting more consistent feature distributions between the fine-tuning and pre-training domains compared to conventional global scaling and shifting operations.

The contributions of this study include:


\begin{itemize}
    \item We propose the Mixture of Physical Priors Adapter (MoPPA), a novel lightweight adapter that leverages multiple physical equations (heat conduction, wave propagation, and Poisson’s equation) to adaptively transform features in pre-trained models.
    
    \item MoPPA utilizes discrete cosine transforms (DCT/IDCT) to operate in the frequency domain, where we assign learnable coefficients based on frequency values for each physical operator. This adaptation enhances the model's ability to adjust feature representations dynamically across spatial and frequency components.

    \item  We introduce a route regularization to prevent the trivial solution of the path weights. It discourages any early bias toward specific path selections, allowing the model to explore diverse configurations early in training.

\end{itemize}

Extensive experiments on image classification and object detection tasks with various pre-training backbones validate that the proposed MoPPA achieves superior performance by adding comparable trainable parameters, compared with state-of-the-art PEFT methods.
Beyond supervised pre-trained models, we also apply MoPPA to fine-tuning on self-supervised models, with results indicating that our approach adapts more effectively across diverse scenarios. 
Besides, adaptation analysis and ablation studies are conducted to verify the effectiveness of MoPPA and the exploration provided by the route regularization.

\section{Related Work}

\subsection{Physics Inspired Models}

\noindent Physical and biological principles have long inspired the development of machine learning models. For instance, the Boltzmann Machine~\citep{boltzmann}, grounded in the Ising model~\citep{ising}, and Hopfield Networks~\citep{hopfield} both utilize energy minimization and probabilistic inference, demonstrating the power of physics-informed approaches in enhancing machine learning.  
Diffusion models~\citep{song2020denoising, ho2020denoising, saharia2022photorealistic}, draw inspiration from Nonequilibrium thermodynamics~\citep{de2013non}, by using Markov chains to model the diffusion process for image generation.  
Physics-Informed Neural Networks (PINNs)~\citep{pinn,piml,scientific,fluid} embed physical laws via PDEs into the neural network learning process, enhancing generalization and interpretability in scientific domains like fluid dynamics.  
Spiking Neural Networks (SNNs)~\citep{ghosh2009spiking, tavanaei2019deep, lee2016training} more accurately replicate the information transmission mechanisms of biological neurons, making them effective tools for simple visual tasks~\citep{bawane2018object}.  
The success of biologically and physically inspired models motivates our exploration of physical priors for adaptive feature alignment and parameter-efficient fine-tuning. Unlike prior physics-informed works, MoPPA uses a lightweight operator to combine multiple physical priors for fine-tuning pre-trained models. 


\subsection{Parameter-Efficient Fine-Tuning}

Early computer vision research primarily focused on enhancing visual representation capabilities by pre-training models on large-scale datasets such as ImageNet-1K~\citep{imagenet,hinton_in1k,resnet}. The pre-training approaches significantly improved performance on various downstream vision tasks~\citep{in1kpt}, demonstrating the effectiveness of extensive labeled data. Recent studies have shifted toward self-supervised pre-training methods, inspired by advancements in natural language processing (NLP)~\citep{bert,gpt4}. These methods achieved outstanding performance across vision tasks, showcasing impressive scalability and enhancing model representation capabilities~\citep{beit,mae,itpn,eva}. As model size increases, however, the costs of full fine-tuning become excessive, which drives the community to explore Parameter-Efficient Fine-Tuning (PEFT) techniques. Unlike full fine-tuning, which updates all model parameters and incurs high computational costs, PEFT aims to maintain competitive performance while reducing the number of trainable parameters and mitigating overfitting risks.

The strategies for PEFT can be coarsely categorized into four: selective parameter updating, adapter-based methods, prompt tuning, and feature transformation. 
Selective parameter updating methods, such as SpotTune~\citep{guo2019spottune} and BitFit~\citep{bitfit}, updated specific layers or bias terms to minimize the number of trainable parameters.
Adapter-based methods, such as Adapter~\citep{adapter}, LoRA~\citep{lora}, AdaptFormer~\citep{adaptformer}, and ARC~\citep{dong2024efficient}, used lightweight modules, low-rank decomposition, and/or parameter sharing across layers for efficient fine-tuning.
Prompt tuning methods, like Visual Prompt Tuning (VPT)~\citep{vpt}, introduced trainable prompts while frozing the backbone during fine-tuning, to take both advantages of prompt learning and lightweight modules.
Feature transformation approaches, such as SSF~\citep{ssf} and FaCT~\citep{jie2023fact}, used scaling, shifting, and decomposition to activate a small proportion of parameters for updating.

While these methods effectively adapt pre-trained models with minimal trainable parameters, they often rely on low-rank priors that may limit flexibility or struggle with generalization across diverse domains with task-specific prompts.

\section{Methodology}

\begin{figure}
    \centering
    \includegraphics[width=0.98\linewidth]{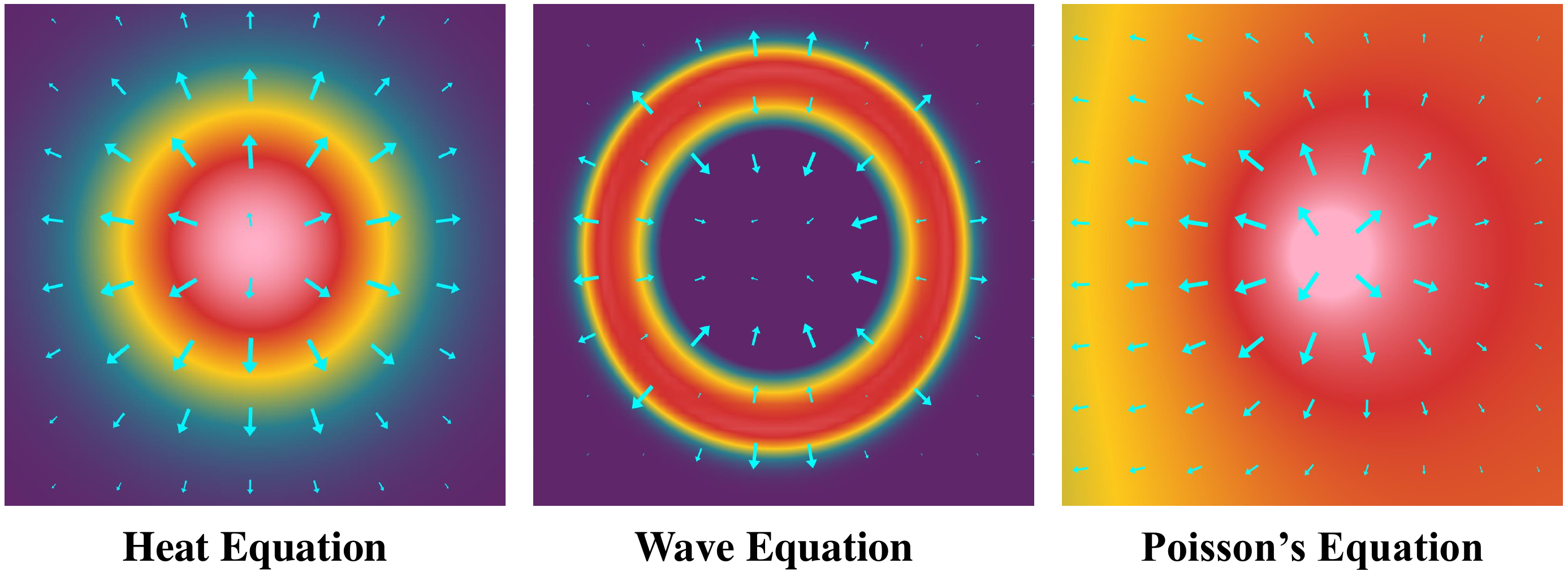}
    \caption{Visualization of diffusion processes of three physical equations. \textbf{Left:} Heat conduction from a central source. \textbf{Mid:} Wave propagation from an initial disturbance. \textbf{Right:} Potential field generated by Poisson's equation with a Dirac delta source in a half-space. More intense colors indicate higher temperatures, higher wave amplitudes, and higher potential values, respectively. }
    \label{fig:phy_illu}
\end{figure}

To introduce the MoPPA method comprehensively, we begin with a review of the three core physical equations it leverages: the Heat Equation, the Wave Equation, and Poisson's Equation. Each of these equations models a distinct type of physical process, visualized in Fig.~\ref{fig:phy_illu}, and can help capture dynamic interactions in our method. By transforming these equations, MoPPA simulates key aspects of their dynamics, enabling a more adaptive and parameter-efficient tuning process for pre-trained models.

\subsection{Preliminaries: Physical Priors}

We will primarily present the formulations of the three functions leveraged by our method, along with their solutions to each respective partial differential equation (PDE). Please refer to Sec.~\ref{sec:derivation} in the supplementary for detailed derivations of these solutions.

\subsubsection{Heat Equation}
Let us define function \( u_H(x, y, t) \):  $\mathbb{R}^2 \times \mathbb{I} \rightarrow \mathbb{R}$, where it represents the temperature at a two-dimensional spatial point $ (x, y)  \in \mathbb{R}^2$  at time  $t \in \mathbb{I}$, $\mathbb{I} \in \mathbb{R}$. The Heat Equation describes how temperature evolves spatially and temporally and can be written as:

%
\begin{equation}\label{eq:heat}
    \frac{\partial{u_H}}{\partial{t}}=k\left(\frac{\partial^2{u_H}}{\partial{x^2}}+\frac{\partial^2{u_H}}{\partial{y^2}}\right),
\end{equation}
where \( k>0 \) is the thermal diffusivity~\citep{heatk}, which measures the rate of heat transfer in a material.
 
Setting the initial condition \( u_H(x,y,t) \big|_{t=0} = u^0_H(x,y) \), the general solution at every time $t$ of the heat equation can be obtained by applying the (inverse) Fourier Transform \( \mathcal{F}(\cdot)/\mathcal{F}^{-1}(\cdot) \) to Eq.~\eqref{eq:solution_freq}, yielding:

\begin{align}\label{eq:solution_spatial_main}
    u_H(x,y,t) &= \mathcal{F}^{-1} \left( \widetilde{u^0_H}(\omega_x, \omega_y) e^{-k(\omega_x^2 + \omega_y^2)t} \right),
\end{align}
where $(\omega_x,\omega_y)$ denotes the coordinate in the frequency domain and $\widetilde{u^0_H}(\omega_x,\omega_y)$ represents the FT-transformed $u^0_H(x,y)$.

\subsubsection{Wave Equation}

Let \( u_W(x, y, t) \) represent the displacement at the point \( (x, y) \) at time \( t \). The classical 2D Wave Equation~\citep{waveeq} can be formulated as:

\begin{equation}\label{eq:wave}
    \frac{\partial^2 u_W}{\partial t^2} = c^2 \left( \frac{\partial^2 u_W}{\partial x^2} + \frac{\partial^2 u_W}{\partial y^2} \right),
\end{equation}
where \(c\) represents the propagation speed of the wave.

We set the initial condition \( u_W(x,y,0) = u^0_W(x,y) \). Besides, to simplify the solution for MoPPA's implementation, we set \( \frac{\partial u_W}{\partial t} \big|_{t=0} = 0 \), which is a common assumption of Neumann boundary condition~\citep{boundary}. By applying the (inverse) Fourier Transform \( \mathcal{F}(\cdot)/\mathcal{F}^{-1}(\cdot) \), the general solution at every time $t$ of the wave equation can be expressed as follows:

\begin{align}\label{eq:solution_spatial_wave_main}
    u_W(x,y,t) &= \mathcal{F}^{-1} \left( \widetilde{u^0_W}(\omega_x, \omega_y) \cos(c \sqrt{\omega_x^2 + \omega_y^2} t) \right),
\end{align}
where $\widetilde{u^0_W}(\omega_x,\omega_y)$ represents the FT-transformed $u^0_W(x,y)$.

\subsubsection{Poisson's Equation}

Let \( u_P(x, y) \) represent a scalar potential function within a two-dimensional region \( D \subset \mathbb{R}^2 \). The classical 2D Poisson's Equation~\citep{possioneq} is defined as:

\begin{equation}\label{eq:poisson}
    \frac{\partial^2 u_P}{\partial x^2} + \frac{\partial^2 u_P}{\partial y^2} = f(x, y),
\end{equation}

where \( f(x, y) \) is a known source term, characterizing the distribution of sources (positive values) or sinks (negative values) within the domain \( D \). Physically, \( f(x, y) \) and \( u_P(x, y) \) take on different interpretations depending on the application. For example, in electrostatics, \( f(x, y) \) represents the charge density distribution, while \( u_P(x, y) \) corresponds to the electric potential. 

Similar to the above, the solution can be obtained by applying the Fourier Transform \( \mathcal{F}(\cdot) \), expressed as follows:

\begin{equation}\label{eq:poisson_solution_main}
    u_P(x, y) = \mathcal{F}^{-1} \left( \frac{-\widetilde{f}(\omega_x, \omega_y)}{\omega_x^2 + \omega_y^2} \right),
\end{equation}
where $\widetilde{f}(\omega_x,\omega_y)$ represents the FT-transformed $f(x,y)$.

\subsection{MoPPA}
\label{sec:moppa}

\begin{figure}[t]
    \centering
    \includegraphics[width=0.99\linewidth]{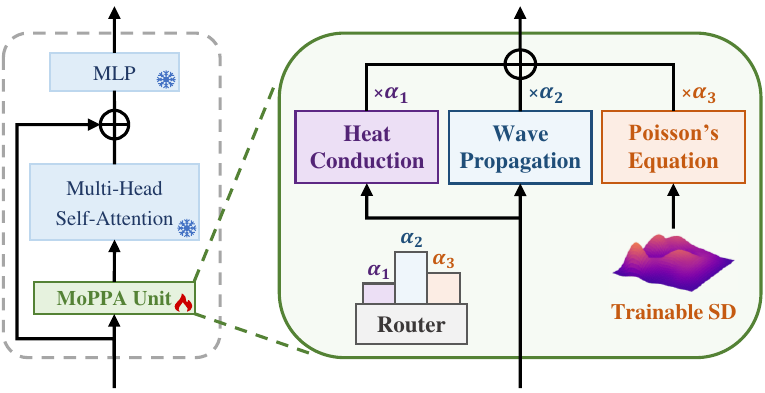}
    \caption{Architecture of a Vision Transformer (ViT) block with an integrated trainable MoPPA unit during fine-tuning. The trainable SD denotes the trainable Source Distribution, which serves as the input to Poisson's Equation. MLP refers to the Multi-Layer Perceptron. The snowflake and fire icons represent frozen and trainable modules, respectively.}
    \label{fig:MoPPA}
\end{figure}

After obtaining the solutions to those equations, we focus on the efficient integration of our MoPPA into existing pre-trained vision models. MoPPA adapts model features by leveraging the dynamics represented by each physical equation, allowing it to capture nuanced spatial and temporal information in a parameter-efficient way.
As shown in Fig.~\ref{fig:MoPPA}, our MoPPA unit, equipped with trainable parameters, is integrated into each block of an existing model, positioned directly before the Multi-Head Attention module. Within each MoPPA unit, the implementation of physical priors follows the transformation formulas outlined in the Preliminaries, specifically Eqs.~\eqref{eq:solution_spatial_main}, \eqref{eq:solution_spatial_wave_main}, and \eqref{eq:poisson_solution_main}. For Poisson's equation, we incorporate a trainable, randomly initialized Source Distribution (SD) in the frequency domain (denoted as $\text{SD}(\omega_x, \omega_y)$), which enables the model to generate adaptable potential fields. Additionally, a routing mechanism assigns learnable path weights to each MoPPA unit, dynamically blending outputs from these different priors.


Given the spatially constrained nature of visual data, along with the fact that its semantic content typically does not extend beyond the image boundaries, we enforce a Neumann boundary condition~\citep{boundary}, expressed as $\frac{\partial u(x, y, t)}{\partial \mathbf{n}} = 0, \quad \forall (x, y) \in \partial D, \, t \geq 0$, where \( \mathbf{n} \) denotes the normal vector to the boundary \( \partial D \). This boundary condition ensures a zero-gradient at edges, naturally handled by the Discrete Cosine Transform (DCT), which represents data with real-valued coefficients and reduces boundary artifacts. Given these advantages, we choose DCT~\citep{dct} over the Discrete Fourier Transform (DFT).

%

The discrete implementations for simulating heat conduction, wave propagation, and frequency Poisson are denoted as \( \text{Heat}(\cdot) \), \( \text{Wave}(\cdot) \), and \( \text{Poisson}(\cdot) \), respectively. Denoting $\mathbf{X} \in \mathbb{R}^{w \times h \times d}$ as the input to the MoPPA unit, these implementations can be formulated as follows:

\begin{equation}\label{eq:heat_implementation}
    \text{Heat}(\text{X}) = \mathbf{IDCT_{2D}}\left(\mathbf{DCT_{2D}}(\text{X})e^{-k \omega^2 t}\right),
\end{equation}
\begin{equation}\label{eq:wave_implementation}
    \text{Wave}(\text{X}) = \mathbf{IDCT_{2D}}\left(\mathbf{DCT_{2D}}(\text{X})\cos\left(c|\omega|t\right)\right),
\end{equation}
\begin{equation}\label{eq:Poisson_implementation}
    \text{Poisson}(\text{X}) = \mathbf{IDCT_{2D}}\left(\frac{\text{SD}(\omega)}{\omega^2+\eta}\right),
\end{equation}
where \( \omega := (\omega_x, \omega_y), \omega^2=\omega_x^2 + \omega_y^2, |\omega|=\sqrt{\omega^2} \), and \( \eta \) is a small constant to ensure numerical stability during division. In all experiments, we set $\eta=0.001$.

Since the output of the \( \mathbf{DCT_{2D}} \) is in the frequency domain, we assign different values for \( k \) and \( c \) in Eq.~\eqref{eq:heat_implementation} and Eq.~\eqref{eq:wave_implementation} as learnable parameters tailored for different frequency values. To limit parameter growth, we split feature channels into multiple heads, similar to multi-head self-attention, and assign shared \( k \) and \( c \) per head. This structure reduces the parameter count by sharing parameters across heads, 
%
%
resulting in \( k := k(\omega,n_i) \) and \( c := c(\omega,n_i) \). We also introduce learnable \( t \) values for \( \text{Heat}(\cdot) \) and \( \text{Wave}(\cdot) \) for each channel dimension \( d_i \) across all heads, yielding \( t := t(d_i) \). For \( \text{Poisson}(\cdot) \), we adopt a similar strategy to reduce the parameters of the trainable Source Distribution \( \text{SD}(\omega) \) in Eq.~\eqref{eq:Poisson_implementation}. We utilize a trainable parameter \( H_1(\omega,n_i) \) for each head \( n_i \) and \( H_2(d_i) \) for each channel dimension \( d_i \). Assuming \( L \) represents the number of feature tokens in self-attention, \( D \) is the number of feature channels, and \( N \) denotes the number of heads, the number of trainable parameters in \( \text{Poisson}(\cdot) \) is reduced from \( LD \) to \( (LN + \frac{D}{N}) \), resulting in a significant reduction. 

\begin{figure}
    \centering
    \includegraphics[width=0.73\linewidth]{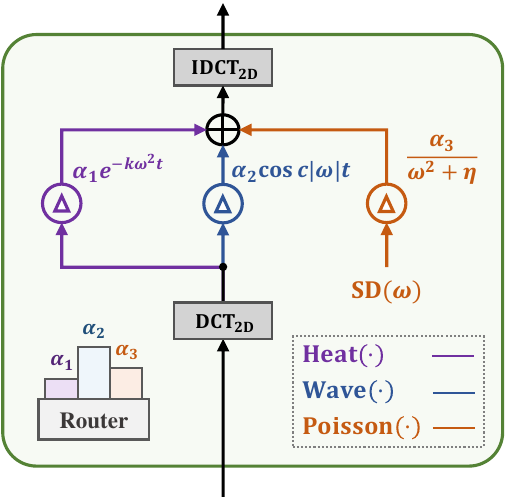}
    \caption{The detailed implementation of a MoPPA unit as described in Eq.~\eqref{eq:moppa_unit}. As shown in the lower right portion, the arrows in \textcolor{purple}{purple}/\textcolor{blue1}{blue}/\textcolor{orange}{orange} represent $\textcolor{purple}{\text{Heat}(\cdot)}/\textcolor{blue1}{\text{Wave}(\cdot)}/\textcolor{orange}{\text{Poisson}(\cdot)}$, respectively.}
    \label{fig:MoPPA2}
\end{figure}

Upon receiving outputs from the three operators, the resulting mixture output \( \text{Y} \) from the MoPPA unit can be expressed as:

\begin{equation}\label{eq:MoPPA_output}
    \text{Y} = \alpha_1\text{Heat}(\text{X}) + \alpha_2\text{Wave}(\text{X}) + \alpha_3\text{Poisson}(\text{X}),
\end{equation}
where \( \alpha_{1,2,3} \) are the coefficients corresponding to the outputs of heat conduction, wave propagation, and frequency Poisson, respectively, and are generated using a softmax function applied to the router's learnable path weights \( \lambda_{i} \).

Considering that the calculations performed by these operators are linear with respect to the input \( \text{X} \), Eq.~\eqref{eq:MoPPA_output} can be implemented as follows:

\begin{equation}
\label{eq:moppa_unit}
    \begin{aligned}
    \text{Y} = \mathbf{IDCT_{2D}}(\mathbf{DCT_{2D}}(\text{X})(\alpha_1 e^{-k \omega^2 t} \\+ \alpha_2 \cos(c|\omega|t)) + \alpha_3 \frac{\text{SD}(\omega)}{\omega^2+\eta}).
    \end{aligned}
\end{equation}

Additionally, Fig.~\ref{fig:MoPPA2} illustrates the calculation process within a MoPPA unit.

\begin{table*}[t]
\centering
\caption{Classification results on VTAB-1K. ``Trainable Params" denotes the average number of trainable parameters across tasks, including the backbone, prompt tokens, and task heads. The number after each domain (Natural, Specialized, Structured) indicates its task count. ``Weighted Average" refers to the average Top-1 accuracy (\%) on VTAB-1K, where the accuracy for each domain is weighted by the number of tasks within that domain. ``IN1K MAE", ``IN1K MOCO v3", and ``IN22K SUP" indicate pre-training with MAE, MOCO v3 on ImageNet-1K, and supervised pre-training on ImageNet-22K with AugReg~\citep{augreg}, respectively.}
\label{table:vtab}
\begin{center}
\vspace{-5pt}
\scalebox{0.96}{
\begin{tabular}{llccccc}
\toprule
\multicolumn{1}{p{2cm}}{\centering Pre-training \\ Methods} & \multicolumn{1}{p{2cm}}{\centering PEFT \\ Methods} &\multicolumn{1}{p{1.5cm}}{\centering Trainable \\ Params} &\multirow{2}{*}{Natural (7)} &\multirow{2}{*}{Specialized (4)} &\multirow{2}{*}{Structured (8)} &\multicolumn{1}{p{1.5cm}}{\centering Weighted \\ Average}\\
\midrule
\multirow{6}{*}{IN1K MAE} & Full & 85.80M & 59.3 & 79.7 & 53.8 & 61.3 \\
&VPT-Deep~\citep{vpt} & 0.60M &36.0 &60.6 &26.6 &37.2 \\
&GateVPT~\citep{gatevpt} & 0.12M & 47.6 & 76.9 & 36.8 & 49.2 \\
&LoRA~\citep{lora} & 0.29M & 57.5 & 77.7 & 57.7 & 61.8 \\
&SPT-LoRA~\citep{sptlora} & 0.38M & 65.4 & 82.4 & 61.5 & 67.3 \\
&SPT-Deep~\citep{spt} & 0.22M & 67.2 &83.2 & 59.2 & 67.2\\
\rblue
&\textbf{MoPPA (Ours)} & 0.26M & \textbf{68.7} & \textbf{84.1} & \textbf{62.7} & \textbf{69.4}\\
\midrule
\multirow{6}{*}{IN1K MOCO v3} & Full & 85.80M & 71.9 & 84.7 & 52.0 & 66.2 \\
&VPT-Deep~\citep{vpt} & 0.60M & 70.3 & 83.0 & 42.4 & 61.2 \\
&GateVPT~\citep{gatevpt} & 0.12M & 74.8 & 83.4 & 49.1 & 65.8 \\
&SPT-Deep~\citep{spt} & 0.22M & 76.2 & 84.9 & 58.4 & 70.5\\
\rblue
&\textbf{MoPPA (Ours)} & 0.26M & \textbf{76.8} & \textbf{85.3} & \textbf{62.0} & \textbf{72.4}\\
\midrule
\multirow{7}{*}{IN22K SUP} & Full & 85.80M & 75.9 & 83.4 & 47.6 & 65.6 \\
&VPT-Deep~\citep{vpt} & 0.60M &78.5 & 82.4 & 55.0 & 69.4 \\
&LoRA~\citep{lora} & 0.29M & 79.5 & 84.6 & 59.8 & 72.3 \\
&SSF~\citep{ssf} & 0.24M & 81.6 & 86.6 & 59.0 & 73.1 \\
&SPT-LoRA~\citep{sptlora} & 0.38M & 81.9 & 85.9 & 61.3 & 74.1 \\
&RLRR~\citep{rlrr} & 0.33M & 83.7 & \textbf{87.3} & 61.5 & 75.1 \\
\rblue
&\textbf{MoPPA (Ours)} & 0.26M & \textbf{85.0} & \textbf{87.3} & \textbf{62.9} & \textbf{76.2}\\
\bottomrule
\end{tabular}
}
\vspace{-5pt}
\end{center}
\end{table*}

\subsection{Route Regularization}

During the initial training phase, the learnable path weights in the router may converge to a trivial solution, such as deactivating two of the paths, which limits the model's ability to explore a range of adaptation strategies and can lead to suboptimal performance. This convergence restricts the model's adaptability, reducing the overall effectiveness of the fine-tuning process.
To address this issue and promote exploration, we introduce a route regularization term that specifically penalizes the concentration of path weights on a single choice. This regularization term is calculated as follows:

\begin{equation}\label{eq:route_regularization}
    \begin{aligned}
        \mathcal{L}_{reg} = \sum_i \alpha_i \log \alpha_i,
    \end{aligned}
\end{equation}
where  $\alpha_i = e^{\lambda_i}/\sum_j e^{\lambda_j}, \quad i=1,2,3$, and $\lambda_i$ denotes the router's learnable path weights.

However, directly incorporating route regularization into the training loss could inadvertently destabilize the optimization process, leading to divergence and degraded performance. To mitigate this, we introduce an adaptive weighting scheme that adjusts based on the training epoch. This adaptive approach allows us to gradually modulate the influence of the regularization term, ensuring it does not interfere with the primary training objective.
Specifically, letting \( T \) denote the current training epoch and \( T_{total} \) represent the total number of epochs, we define the final training loss \( \mathcal{L} \) as follows:

\begin{equation}\label{eq:final_loss}
    \mathcal{L} = \mathcal{L}_{origin} + w \max\left(1 - \frac{2T}{T_{total}}, 0\right) \mathcal{L}_{reg},
\end{equation}
where \( \mathcal{L}_{origin} \) denotes the original training loss and \( w \) serves as a coefficient that balances the contributions of both loss terms.  As a result, in the early stages of training, the route regularization encourages the learnable path weights to explore a wider range of options, fostering a more robust optimization landscape. As training progresses, the influence of the route regularization diminishes adaptively, thereby helping to maintain the stability and integrity of the optimization objective while still facilitating exploration during the initial stage.

\section{Experiment}

\subsection{Setting}

\noindent\textbf{Datasets.} To validate the effectiveness and generalization of MoPPA, we conduct experiments across various vision tasks, including image classification, object detection, and out-of-distribution classification. The evaluation datasets are as follows:

\textit{VTAB-1K}~\citep{vtab} comprises 19 tasks from diverse domains, featuring \textit{natural} images from standard cameras, \textit{specialized} images from non-standard sources (such as remote sensing and medical cameras), and \textit{structured} images from simulated environments. We utilize the 800-200 train/val split as established in previous works~\citep{vpt,ssf}.

\textit{FGVC} includes five fine-grained classification datasets: CUB-200-2011~\citep{cub}, NABirds~\citep{nabirds}, Oxford Flowers~\citep{flowers}, Stanford Dogs~\citep{dogs}, and Stanford Cars~\citep{cars}. Following VPT~\citep{vpt}, we randomly split the training set into 90\% for training and 10\% for validation.

\textit{ImageNet-1K}~\citep{imagenet} is a large-scale image classification dataset with 1,000 classes, containing over 1M images. 

\textit{MS COCO}~\citep{coco} is a widely-used large-scale dataset for evaluating object detection and instance segmentation.

\noindent\textbf{Pre-trained Models.} To ensure a fair and comprehensive comparison, we utilize the ImageNet-1K MAE~\citep{mae} pre-trained, ImageNet-1K MOCO v3~\citep{MOCOv3} pre-trained, and ImageNet-22K pre-trained ViT-B/16~\citep{vit} as baseline models for image classification tasks. Additionally, we select the ImageNet-22K pre-trained Swin-B~\citep{swin} as the baseline for object detection and instance segmentation tasks. We also evaluated MoPPA with pre-trained ViT-Large, and please refer to Sec.~\ref{sec:vit_large} in the supplementary for detailed results with pre-trained ViT-L.

\begin{table*}[ht]\centering
\caption{Classification results on FGVC. The term ``Trainable Params" refers to the average count of trainable parameters across all tasks, encompassing the backbone, prompt tokens, and task heads. ``IN1K MAE" indicates that the model is pre-trained by MAE on ImageNet-1K, while ``IN22K SUP" signifies that the model undergoes supervised pre-training on ImageNet-22K without AugReg~\citep{augreg}, respectively. }
\label{table:fgvc}
\begin{center}
\vspace{-5pt}
\scalebox{0.85}{
\begin{tabular}{llccccccc}
\toprule
\multicolumn{1}{p{1.7cm}}{\centering Pre-training \\ Methods} & \multicolumn{1}{p{1.5cm}}{\centering PEFT \\ Methods} &\multicolumn{1}{p{1.4cm}}{\centering Trainable \\ Params} &\multirow{2}{*}{CUB-200-2011} &\multirow{2}{*}{NABirds} &\multirow{2}{*}{ Oxford Flowers} &\multirow{2}{*}{Stanford Dogs} &\multirow{2}{*}{Stanford Cars} &\multirow{2}{*}{Average} \\
\midrule
\multirow{6}{*}{IN1K MAE} & Full & 85.98M & \textbf{80.6} & \textbf{77.9} & 91.7 & 80.4 & 83.5 & 82.8 \\
&VPT-Deep~\citep{vpt} & 0.85M & 68.3 & 65.2 & 80.1 & 78.8 & 67.7 & 72.0 \\
&GateVPT~\citep{gatevpt} & 0.27M & 70.6 & 67.3 & 78.6 & 78.9 & 71.7 & 73.4 \\
&SPT-Deep~\citep{spt} & 0.37M & 80.1 & 76.3 & 93.1 & 82.2 & 84.6 & 83.3 \\
\rblue
&\textbf{MoPPA (Ours)} & 0.40M & \textbf{80.6} & 77.0 & \textbf{93.5} & \textbf{82.4} & \textbf{86.8} & \textbf{84.1} \\
\midrule
\multirow{7}{*}{IN22K SUP} & Full & 85.98M & 87.3 & 82.7 & 98.8 & 89.4 & 84.5 & 88.5 \\
&VPT-Deep~\citep{vpt} & 0.85M & 88.5 & 84.2 & 99.0 & 90.2 & 83.6 & 89.1 \\
&LORA~\citep{lora} & 0.44M & 88.3 & \textbf{85.6} & 99.2 & 91.0 & 83.2 & 89.5 \\
&AdaptFormer~\citep{adaptformer} & 0.46M & 88.4	& 84.7 & 99.2 & 88.2 & 81.9 & 88.5 \\
&RLRR~\citep{rlrr} & 0.47M & 89.3 & 84.7 & 99.5 & 92.0 & 87.0 & 90.4 \\
\rblue
&\textbf{MoPPA (Ours)} & 0.40M & \textbf{89.4} & 85.1 & \textbf{99.6} & \textbf{92.2} & \textbf{88.5} & \textbf{91.0} \\
\bottomrule
\end{tabular}}
\end{center}
\vspace{-5pt}
\end{table*}

\noindent\textbf{Implementation Details.} During the training phase, we apply standard data augmentation techniques as described in VPT~\citep{vpt}. For the five FGVC datasets, we employ random horizontal flips and randomly resize crops to 224$\times$224 resolution. In the case of the VTAB-1K benchmark, images are resized to 224$\times$224 resolution, accompanied by random horizontal flips across all 19 datasets. We insert MoPPA units before self-attention operators in ViT and Swin on classification and object detection tasks. Additionally, given the relatively few training parameters of MoPPA units, we also incorporate global \& scaling / convolution operations on classification / detection tasks for PEFT to ensure that the total number of training parameters is comparable to that of the baseline methods for fair comparison. Please refer to Sec.~\ref{sec:detailed_settings} in the supplementary for training details. All experiments are executed using PyTorch 2.2 tools~\citep{pytorch} on NVIDIA 40GB A100 GPUs.

\subsection{Performance}

\noindent\textbf{VTAB-1K.} Table~\ref{table:vtab} summarizes the results on VTAB-1K. Our proposed MoPPA consistently outperforms other PEFT methods across diverse classification tasks, achieving a leading weighted average accuracy of \textbf{69.4\%} with only \textbf{0.26M} trainable parameters in the ImageNet-1K MAE pre-training scenario. MoPPA also excels across all three domains: \textbf{68.7\%} for Natural, \textbf{84.1\%} for Specialized, and \textbf{62.7\%} for Structured tasks.
Compared to low-rank methods, MoPPA achieves 7.6\% and 2.1\% higher weighted accuracy than LoRA (61.8\%) and SPT-LoRA (67.3\%) with fewer parameters, highlighting the advantage of physical priors over low-rank priors in PEFT. MoPPA also outperforms prompt tuning methods, with improvements of 32.2\% and 20.2\% over VPT-Deep and GateVPT, respectively, reinforcing the superiority of physical priors over learnable visual prompts. 
When using a MOCO v3 pre-trained ViT-B backbone, MoPPA achieves \textbf{72.4\%}, surpassing SPT-Deep by 1.9\% and full fine-tuning by 6.2\%. In the ImageNet-22K supervised setting, MoPPA achieves \textbf{76.2\%}, outperforming RLRR and full fine-tuning by 1.1\% and 10.6\%, respectively. These results highlight the adaptability and robustness of MoPPA across different pre-training paradigms. 
By modeling feature transformations through well-established physical equations, MoPPA not only achieves competitive performance but also enhances robustness and stability, making it a highly effective and parameter-efficient approach. Please refer to Sec.~\ref{sec:vtab_per_task} for VTAB-1K per-task results in the supplementary material.

\noindent\textbf{FGVC.} Table~\ref{table:fgvc} compares the classification results of MoPPA with various PEFT methods on FGVC under two pre-training scenarios: ImageNet-1K MAE and ImageNet-22K supervised. In the ImageNet-1K MAE pre-training setting, MoPPA achieves the highest average Top-1 accuracy of \textbf{84.1\%} with only \textbf{0.40M} trainable parameters, outperforming Full Fine-Tuning and PEFT baselines such as VPT-Deep, GateVPT, and SPT-Deep. 
In the ImageNet-22K supervised pre-training scenario, MoPPA achieves the best average accuracy of \textbf{91.0\%}, surpassing methods like RLRR and AdaptFormer. These results highlight MoPPA's adaptability and robustness across FGVC tasks and pre-training paradigms. The incorporation of physical priors effectively enhances fine-tuning performance, validating MoPPA’s strength in fine-grained image classification.

\begin{table}[t]
\centering
\caption{Image classification results on ImageNet-1K with ImageNet-22K pre-trained ViT-B backbone (with AugReg~\citep{augreg}).}
\label{table:in1k_ft}
\vspace{-5pt}
\begin{tabular}{lcc}
\toprule
PEFT Methods & Trainable Params & top-1 acc. (\%) \\
\midrule
Full Fine-tuning & 86.57M & 83.6 \\
Linear probing & 0.77M & 82.0 \\
Adapter~\citep{adapter} & 1.00M & 82.7 \\
VPT-Deep~\citep{vpt} & 1.23M & 82.5 \\
SSF~\citep{vpt} & 0.97M & 83.1 \\
\rblue
\textbf{MoPPA (Ours)} & 0.99M & \textbf{83.9} \\
\bottomrule
\end{tabular}
\vspace{-5pt}
\end{table}

\noindent\textbf{ImageNet-1K.} Table~\ref{table:in1k_ft} presents the classification results of various PEFT methods on ImageNet-1K using the ImageNet-22K pre-trained ViT-B backbone. Our proposed MoPPA achieves a Top-1 accuracy of \textbf{83.9\%} with only \textbf{0.99M} trainable parameters, outperforming all compared methods, including SSF (83.1\%), VPT-Deep (82.5\%), and Adapter (82.7\%). These results validate the effectiveness of MoPPA in leveraging physical priors for PEFT, particularly when abundant training samples are available.
%

\begin{table}[t]
\centering
\caption{Object detection and instance segmentation results on COCO~\citep{coco} with ImageNet-22K pre-trained Swin-B~\citep{swin} backbone. All PEFT methods utilize the Cascade Mask R-CNN as the detector for a fair comparison. AP$^b$ and AP$^m$ represent box AP and mask AP, respectively.}
\label{table:coco_swin}
\vspace{-5pt}
\begin{tabular}{lccc}
\toprule
PEFT Methods & Trainable Prams & AP$^b$ & AP$^m$ \\
\midrule
Full Fine-tuning & 89.14M & 52.4 & 45.1 \\
Partial-1~\citep{partial1} & 12.95M & 50.6 & 43.7 \\
Adapter~\citep{adapter} & 3.19M & 52.1 & 45.0 \\
LoRA~\citep{lora} & 3.06M & 50.4 & 43.9 \\
LoRand~\citep{adaptformer} & 4.68M & 51.9 & 44.7 \\
\rblue
\textbf{MoPPA (Ours)} & 3.18M & \textbf{52.7} & \textbf{45.6} \\
\bottomrule
\end{tabular}
\vspace{-5pt}
\end{table}

\noindent\textbf{Detection \& Segmentation.} To assess MoPPA's performance on downstream tasks, we evaluated it on the COCO benchmark~\citep{coco} using an ImageNet-22K pre-trained Swin-B backbone and Cascade Mask R-CNN~\citep{cascade} (36 epochs). MoPPA achieves \textbf{52.7 / 45.6} Box and Mask APs with only \textbf{3.18M} trainable parameters, as shown in Table~\ref{table:coco_swin}, outperforming Partial-1 and LoRA, and slightly surpassing Adapter and LoRand. These results demonstrate the benefits of physical priors and MoPPA’s strong generalization across vision tasks.

\subsection{Adaptation Analysis}
\label{sec:analysis_main}

To investigate MoPPA's adaptation capacity, we compare it against LoRA by training both to regress randomly generated input tensors to their corresponding randomly generated Ground Truth (GT) tensors. Pre-trained ViT-B models equipped with LoRA ($\text{rank}=6$ for parameter alignment) and MoPPA are fine-tuned under identical settings, using Mean Squared Error (MSE) as the supervision metric. Please refer to Sec.~\ref{sec:analysis_implementation} for detailed implementation and results. Across 5 trials, LoRA/MoPPA achieved MSEs of ${0.065}\pm{0.0007} / {0.050}\pm{0.0011}$, respectively. Fig.~\ref{fig:moppa_error} visualizes the GT and absolute errors of LoRA / MoPPA predictions in one trial. MoPPA's predictions are closer to GT compared with LoRA, highlighting its effective use of physical priors, which yield more accurate adaptations compared to low-rank priors in LoRA.


\begin{figure}[t]
    \centering
    \includegraphics[width=0.98\linewidth]{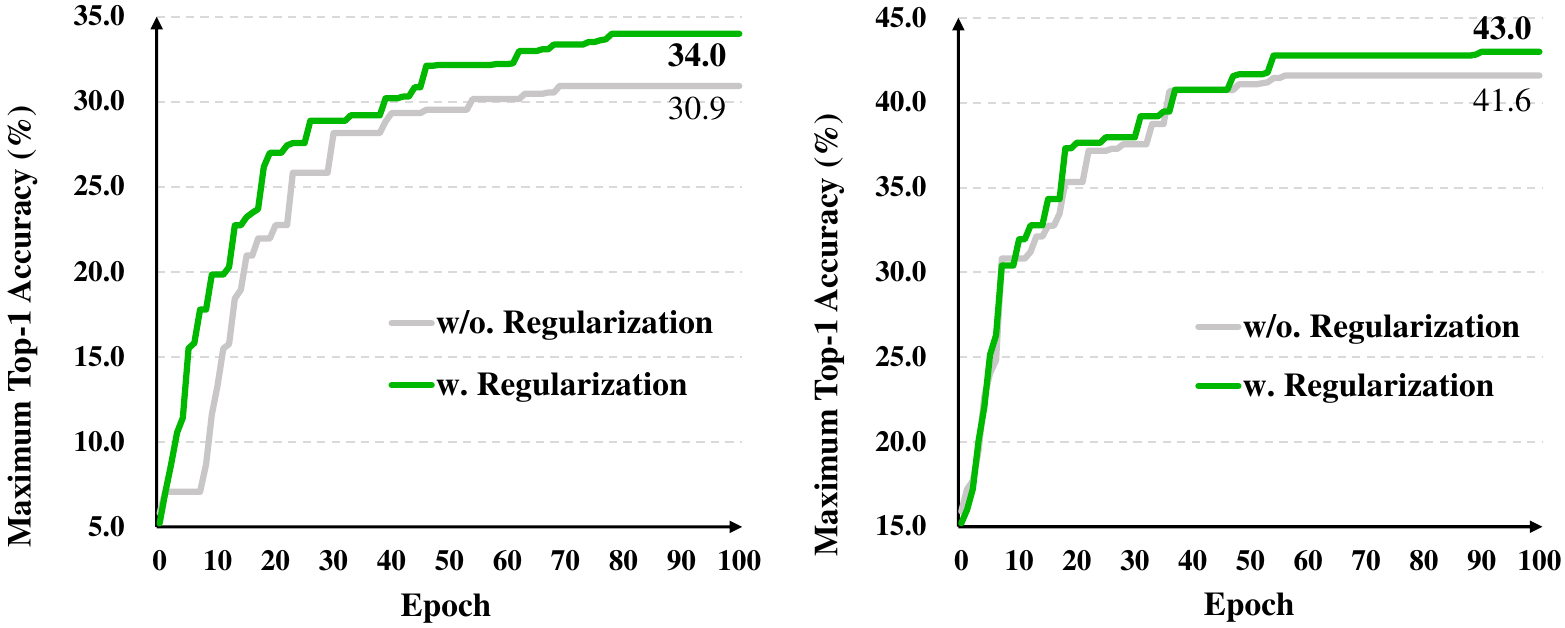}
    \vspace{-5pt}
    \caption{Maximum Top-1 Accuracy (\%) curves over epochs on Smallnorb-Azi (\textbf{Left}) and Smallnorb-Ele (\textbf{Right}) in VTAB-1K, where ``w. Regularization" and ``w/o. Regularization" denote training with and without the route regularization, respectively.}
    \label{fig:ablation_route}
\end{figure}

\begin{table}
\caption{Ablation study of physical priors on VTAB-1K using an ImageNet-22K pre-trained ViT-B/16 backbone. ``Trainable Params" excludes the classification head.}
\label{table:ablation_priors}
\centering
\vspace{-5pt}
\begin{tabular}{lcc}
    \toprule
    Settings & Trainable Params & top-1 acc. (\%) \\ 
    \midrule
    MoPPA & 0.225M & 76.2 \\
    w/o $\text{Poisson}(\cdot)$ & 0.224M & 75.3 \\
    w/o $\text{Wave}(\cdot)$ & 0.224M & 75.2 \\
    w/o $\text{Heat}(\cdot)$ & 0.224M & 74.9 \\
    w/o MoPPA units & 0.222M & 74.2 \\
    \bottomrule
\end{tabular}
\vspace{-5pt}
\end{table}

\subsection{Ablation Studies}

\noindent\textbf{Physical Priors.} To assess the impact of each physical prior, we evaluated MoPPA on VTAB-1K by individually removing $\text{Heat}(\cdot)$, $\text{Wave}(\cdot)$, and $\text{Poisson}(\cdot)$. As shown in Table~\ref{table:ablation_priors}, the absence of any prior leads to a significant performance drop, confirming the contribution of each component. Furthermore, removing all MoPPA units results in a 2.0\% decrease of VTAB-1K Top-1 Accuracy, highlighting the overall effectiveness of MoPPA.

\noindent\textbf{The route regularization.} To validate the effectiveness of our proposed route regularization, we tested MoPPA with and without the route regularization with ImageNet-22K pre-trained ViT backbone on Smallnorb-Azi and Smallnorb-Ele in VTAB-1K. Maximum Top-1 Accuracy curves in Fig.~\ref{fig:ablation_route} illustrate that the route regularization effectively helps the optimization process of MoPPA, enabling it to achieve better performance.

\section{Conclusion}

We have introduced MoPPA, a lightweight visual operator designed for parameter-efficient fine-tuning (PEFT) of vision models. MoPPA leverages physical priors by integrating Heat Diffusion, Wave Propagation, and Poisson’s Equation to create adaptable structures that dynamically adjust based on local and global contexts. Our route regularization mechanism ensures these priors work synergistically, enabling optimal performance across diverse tasks. Extensive experiments demonstrate that MoPPA outperforms existing PEFT methods, providing superior accuracy with comparable parameter budgets, and offers a practical solution for large model adaptation in visual models. More importantly, exploring diverse physical priors across applications could further enhance representational power.


%
{
    \small
    \bibliographystyle{ieeenat_fullname}
    \bibliography{main}
}

\clearpage
\appendix

\section{Derivation of Three Physical Equations}
\label{sec:derivation}

\subsection{Heat Equation}

Let \( u_H(x, y, t) \) denote the temperature at the point \( (x, y) \) at time \( t \) within a two-dimensional region \( D \subset \mathbb{R}^2 \). The classical heat equation~\citep{heateq} can be expressed as
\begin{equation}\label{eq:heat_supp}
    \frac{\partial{u_H}}{\partial{t}}=k\left(\frac{\partial^2{u_H}}{\partial{x^2}}+\frac{\partial^2{u_H}}{\partial{y^2}}\right),
\end{equation}
where \( k > 0 \) denotes the thermal diffusivity~\citep{heatk}. It measures the rate of heat transfer within a material.
 
Setting the initial condition \( u_H(x,y,t) \big|_{t=0} = f(x,y) \), we derive the general solution at every time $t$ of Eq.~\eqref{eq:heat_supp} by applying the Fourier Transform (denoted as \( \mathcal{F} \)) to both sides of the equation, as

\begin{equation}\label{eq:ft}
    \mathcal{F} \left( \frac{\partial u_H}{\partial t} \right) = k \mathcal{F} \left( \frac{\partial^2 u_H}{\partial x^2} + \frac{\partial^2 u_H}{\partial y^2} \right).
\end{equation}

Let's define \( \widetilde{u_H}(\omega_x, \omega_y, t) \) as the Fourier Transform of \( u_H(x,y,t) \), that is, \( \widetilde{u_H}(\omega_x, \omega_y, t) \coloneqq \mathcal{F}(u_H(x,y,t)) \). Consequently, the left-hand side of Eq.~\eqref{eq:ft} is expressed as

\begin{equation}\label{eq:LHS}
    \mathcal{F} \left( \frac{\partial u_H}{\partial t} \right) = \frac{\partial \widetilde{u_H}(\omega_x, \omega_y, t)}{\partial t}.
\end{equation}

Utilizing the derivative property of the Fourier Transform, the right-hand side of Eq.~\eqref{eq:ft} is transformed to

\begin{equation}\label{eq:RHS}
    \mathcal{F} \left( \frac{\partial^2 u_H}{\partial x^2} + \frac{\partial^2 u_H}{\partial y^2} \right) = -(\omega_x^2 + \omega_y^2) \widetilde{u_H}(\omega_x, \omega_y, t).
\end{equation}

By combining the expressions derived from both sides, we can rewrite Eq.~\eqref{eq:ft} as an ordinary differential equation (ODE) in the frequency domain, as

\begin{equation}\label{eq:ode}
    \frac{d \widetilde{u_H}(\omega_x, \omega_y, t)}{dt} = -k(\omega_x^2 + \omega_y^2) \widetilde{u_H}(\omega_x, \omega_y, t).
\end{equation}

By imposing the initial condition \( \widetilde{u_H}(\omega_x, \omega_y, t) \big|_{t=0} = \widetilde{f}(\omega_x, \omega_y) \) (where \( \widetilde{f}(\omega_x, \omega_y) \) represents the Fourier Transform of \( f(x,y) \)), we can solve for \( \widetilde{u_H}(\omega_x, \omega_y, t) \) in Eq.~\eqref{eq:ode}, as

\begin{equation}\label{eq:solution_freq}
    \widetilde{u_H}(\omega_x, \omega_y, t) = \widetilde{f}(\omega_x, \omega_y) e^{-k(\omega_x^2 + \omega_y^2)t}.
\end{equation}

As a result, the general solution at every time $t$ of the heat equation in the spatial domain can be obtained by applying the inverse Fourier Transform \( \mathcal{F}^{-1} \) to Eq.~\eqref{eq:solution_freq}, as

\begin{align}\label{eq:solution_spatial}
    u_H(x,y,t) &= \mathcal{F}^{-1} \left( \widetilde{f}(\omega_x, \omega_y) e^{-k(\omega_x^2 + \omega_y^2)t} \right).
\end{align}

\subsection{Wave Equation}

Let \( u_W(x, y, t) \) denote the displacement of point \( (x, y) \) at time \( t \) within a two-dimensional domain \( D \subset \mathbb{R}^2 \). The classical wave equation~\citep{waveeq} is formulated as

\begin{equation}\label{eq:wave_supp}
    \frac{\partial^2 u_W}{\partial t^2} = c^2 \left( \frac{\partial^2 u_W}{\partial x^2} + \frac{\partial^2 u_W}{\partial y^2} \right),
\end{equation}
where \(c\) denotes the propagation speed of the wave.

To derive the general solution for \( u_W(x,y,t) \), we set the initial conditions \( u_W(x,y,0) = f(x,y) \). Besides, to simplify the solution for MoPPA's implementation, we set \( \frac{\partial u}{\partial t} \big|_{t=0} = 0 \), which is a common assumption of Neumann boundary condition~\citep{boundary}. By applying the Fourier Transform ($\mathcal{F}$) to both sides of Eq.~\eqref{eq:wave_supp}, we have

\begin{equation}\label{eq:ft_wave}
    \mathcal{F} \left( \frac{\partial^2 u_W}{\partial t^2} \right) = c^2 \mathcal{F} \left( \frac{\partial^2 u_W}{\partial x^2} + \frac{\partial^2 u_W}{\partial y^2} \right).
\end{equation}

Let us denote \( \widetilde{u_W}(\omega_x, \omega_y, t) \) as the Fourier Transform of \( u_W(x,y,t) \), defined as \( \widetilde{u_W}(\omega_x, \omega_y, t) \coloneqq \mathcal{F}(u_W(x,y,t)) \). The left-hand side of Eq.~\eqref{eq:ft_wave} is expressed as

\begin{equation}\label{eq:LHS_wave}
    \mathcal{F} \left( \frac{\partial^2 u_W}{\partial t^2} \right) = \frac{\partial^2 \widetilde{u_W}(\omega_x, \omega_y, t)}{\partial t^2}.
\end{equation}

Utilizing the properties of the Fourier Transform, we can rewrite the right-hand side of Eq.~\eqref{eq:ft_wave} as

\begin{equation}\label{eq:RHS_wave}
    \mathcal{F} \left( \frac{\partial^2 u_W}{\partial x^2} + \frac{\partial^2 u_W}{\partial y^2} \right) = -(\omega_x^2 + \omega_y^2) \widetilde{u_W}(\omega_x, \omega_y, t).
\end{equation}

Combining the expressions from both sides leads us to the ordinary differential equation (ODE) in the frequency domain, as

\begin{equation}\label{eq:ode_wave}
    \frac{d^2 \widetilde{u_W}(\omega_x, \omega_y, t)}{dt^2} + c^2 (\omega_x^2 + \omega_y^2) \widetilde{u_W}(\omega_x, \omega_y, t) = 0.
\end{equation}

This ODE describes a simple harmonic oscillator. The general solution can be expressed in terms of the initial conditions as follows:

\begin{equation}\label{eq:solution_wave_simplified}
    \widetilde{u_W}(\omega_x, \omega_y, t) = \widetilde{f}(\omega_x, \omega_y) \cos(c \sqrt{\omega_x^2 + \omega_y^2} t),
\end{equation}
where \( \widetilde{f}(\omega_x, \omega_y) \) denotes the Fourier Transform of \( f(x,y) \). Finally, to retrieve the solution in the spatial domain at any time \( t \), we apply the inverse Fourier Transform \( \mathcal{F}^{-1} \), as

\begin{align}\label{eq:solution_spatial_wave_supp}
    u_W(x,y,t) &= \mathcal{F}^{-1} \left( \widetilde{f}(\omega_x, \omega_y) \cos(c \sqrt{\omega_x^2 + \omega_y^2} t) \right).
\end{align}

\begin{table*}[h]
	\centering
	\caption{VTAB-1K Per-task results with pre-trained ViT-B. ``IN1K MAE", ``IN1K MOCO v3", and ``IN22K SUP" indicate pre-training with MAE, MOCO v3 on ImageNet-1K, and supervised pre-training on ImageNet-22K, respectively. }
	\resizebox{\linewidth}{!}{
		\begin{tabular}{c|ccccccc|c|cccc|c|cccccccc|c|cc}
			\toprule
			\multirow{2}{*}[-27pt]{\textbf{Methods}} & \multicolumn{8}{c|}{\textbf{Natural}}                         & \multicolumn{5}{c|}{\textbf{Specialized}} & \multicolumn{9}{c|}{\textbf{Structed}}                                &       &  \\
			 & \rotatebox{90}{\textbf{CIFAR-100}} & \rotatebox{90}{\textbf{Caltech101}} & \rotatebox{90}{\textbf{DTD}} & \rotatebox{90}{\textbf{Flowers102}} & \rotatebox{90}{\textbf{Pets}} & \rotatebox{90}{\textbf{SVHN}} & \multicolumn{1}{c}{\rotatebox{90}{\textbf{Sun397}}} & \rotatebox{90}{\textbf{Average}} & \rotatebox{90}{\textbf{Camelyon}} & \rotatebox{90}{\textbf{EuroSAT}} & \rotatebox{90}{\textbf{Resisc45}} & \multicolumn{1}{c}{\rotatebox{90}{\textbf{Retinopathy}}} & \rotatebox{90}{\textbf{Average}} & \rotatebox{90}{\textbf{Clevr-Count}} & \rotatebox{90}{\textbf{Clevr-Dist}} & \rotatebox{90}{\textbf{DMLab}} & \rotatebox{90}{\textbf{KITTI-Dist}} & \rotatebox{90}{\textbf{dSpr-Loc}} & \rotatebox{90}{\textbf{dSpr-Ori}} & \rotatebox{90}{\textbf{sNORB-Azim}} & \multicolumn{1}{c}{\rotatebox{90}{\textbf{sNORB-Ele}}} & \rotatebox{90}{\textbf{Average}} & \rotatebox{90}{\textbf{Average Total}} & \rotatebox{90}{\textbf{Params.(M)}} \\
            \midrule
            \multicolumn{25}{c}{\textit{IN1K MAE}} \\
            \midrule
		  MoPPA & 39.1 & 90.4 & 62.9 & 85.6 & 86.3 & 89.6 & 26.9 & 68.7 & 86.1 & 94.3 & 80.4 & 75.6 & 84.1 & 81.1 & 63.7 & 51.4 & 82.0 & 84.7 & 56.2 & 37.5 & 44.5 & 62.7 & 69.4 & 0.26 \\
		  \midrule
            \multicolumn{25}{c}{\textit{IN1K MOCO v3}} \\
            \midrule
		  MoPPA & 62.6 & 92.5 & 69.4 & 92.1 & 88.3 & 89.6 & 42.8 & 76.8 & 86.9 & 95.5 & 83.6 & 75.2 & 85.3 & 82.5 & 64.5 & 49.1 & 83.2 & 84.8 & 53.6 & 33.1 & 45.4 & 62.0 & 72.4 & 0.26 \\
		  \midrule
            \multicolumn{25}{c}{\textit{IN22K SUP}} \\
            \midrule
		  Full fine-tuning & 68.9  & 87.7  & 64.3  & 97.2  & 86.9  & 87.4  & 38.8  & 75.9  & 79.7  & 95.7  & 84.2  & 73.9  & 83.4  & 56.3  & 58.6  & 41.7  & 65.5  & 57.5  & 46.7  & 25.7  & 29.1  & 47.6  & 65.6  & 85.80  \\
		  VPT-Deep~\cite{vpt} & 78.8 & 90.8  & 65.8  & 98.0  & 88.3  & 78.1  & 49.6  & 78.5  & 81.8  & 96.1  & 83.4  & 68.4  & 82.4  & 68.5  & 60.0  & 46.5  & 72.8  & 73.6  & 47.9  & 32.9  & 37.8  & 55.0  & 69.4  & 0.60  \\
		  LoRA~\cite{lora}  & 67.1  & 91.4 & 69.4  & 98.8  & 90.4  & 85.3  & 54.0  & 79.5  & 84.9 & 95.3  & 84.4  & 73.6  & 84.6  & 82.9 & \textbf{69.2} & 49.8 & 78.5  & 75.7  & 47.1  & 31.0  & \textbf{44.0} & 59.8  & 72.3  & 0.29  \\
            SSF~\cite{ssf}   & 69.0  & 92.6 & 75.1  & 99.4  & 91.8  & 90.2  & 52.9 & 81.6  & 87.4 & 95.9  & 87.4  & 75.5  & 86.6  & 75.9  & 62.3 & 53.3  & 80.6  & 77.3  & 54.9 & 29.5  & 37.9  & 59.0  & 73.1  & 0.24  \\
            SPT-LoRA~\citep{sptlora} & 73.5 & 93.3 & 72.5 & 99.3 & 91.5 & 87.9 & 55.5 & 81.9 & 85.7 & \textbf{96.2} & 85.9 & 75.9 & 85.9 & \textbf{84.4} & 67.6 & 52.5 & 82.0 & 81.0 & 51.1 & 30.2 & 41.3 & 61.3 & 74.1 & 0.38 \\
            RLRR~\citep{rlrr} & 76.7 & 92.7 & 76.3 & 99.6 & \textbf{92.6} & 91.8 & 56.0 & 83.7 & \textbf{87.8} & \textbf{96.2} & 89.1 & 76.3 & \textbf{87.3} & 80.4 & 63.3 & 54.5 & \textbf{83.3} & 83.0 & 53.7  & 32.0 & 41.7 & 61.5 & 75.1 & 0.33 \\
            \rblue
		  \textbf{MoPPA (Ours)} & \textbf{79.7} & \textbf{94.9} & \textbf{78.3} & \textbf{99.7} & 92.4 & \textbf{92.4} & \textbf{57.5} & \textbf{85.0} & 87.6 & 96.1 & \textbf{89.2} & \textbf{76.4} & \textbf{87.3} & 81.4 & 63.7 & \textbf{54.6} & \textbf{83.3} & \textbf{86.7} & \textbf{56.2} & \textbf{34.0} & 43.1 & \textbf{62.9} & \textbf{76.2} & 0.26 \\
            \bottomrule
		\end{tabular}
	}
	\label{table:vtab_per_task}
\end{table*}

\subsection{Poisson's Equation}

Let \( u_P(x, y) \) represent a scalar potential function within a two-dimensional region \( D \subset \mathbb{R}^2 \). The classical 2D Poisson's equation~\citep{possioneq} is defined as:
\begin{equation}\label{eq:poisson_eq}
    \frac{\partial^2 u_P}{\partial x^2} + \frac{\partial^2 u_P}{\partial y^2} = f(x, y),
\end{equation}
where \( f(x, y) \) is a known source term that describes the distribution of sources (positive values) or sinks (negative values) within the domain \( D \). Depending on the context, \( f(x, y) \) and \( u_P(x, y) \) may have various interpretations. For example, in electrostatics, \( f(x, y) \) corresponds to the charge density, while \( u_P(x, y) \) represents the electric potential.

To solve Eq.~\eqref{eq:poisson_eq}, we apply the Fourier Transform \( \mathcal{F} \) to both sides of the equation. Let \( \widetilde{u}_P(\omega_x, \omega_y) \) and \( \widetilde{f}(\omega_x, \omega_y) \) denote the Fourier Transforms of \( u_P(x, y) \) and \( f(x, y) \), respectively. Using the linearity of the Fourier Transform, the equation becomes:
\begin{equation}\label{eq:poisson_ft}
    \mathcal{F} \left( \frac{\partial^2 u_P}{\partial x^2} \right) + \mathcal{F} \left( \frac{\partial^2 u_P}{\partial y^2} \right) = \mathcal{F}(f(x, y)).
\end{equation}

With the derivative property, we obtain:
\begin{equation}\label{eq:poisson_freq}
    -(\omega_x^2 + \omega_y^2) \widetilde{u}_P(\omega_x, \omega_y) = \widetilde{f}(\omega_x, \omega_y).
\end{equation}

Rearranging for \( \widetilde{u}_P(\omega_x, \omega_y) \), we find:
\begin{equation}\label{eq:poisson_solution_freq}
    \widetilde{u}_P(\omega_x, \omega_y) = \frac{-\widetilde{f}(\omega_x, \omega_y)}{\omega_x^2 + \omega_y^2}.
\end{equation}

To obtain the solution in the spatial domain, we apply the inverse Fourier Transform \( \mathcal{F}^{-1} \) to Eq.~\eqref{eq:poisson_solution_freq}:
\begin{equation}\label{eq:poisson_solution_spatial}
    u_P(x, y) = \mathcal{F}^{-1} \left( \frac{-\widetilde{f}(\omega_x, \omega_y)}{\omega_x^2 + \omega_y^2} \right).
\end{equation}

\section{Detailed Training Settings}
\label{sec:detailed_settings}

For classification tasks, we employ AdamW~\citep{adamw} as the optimizer in PEFT. The training schedule includes a warm-up phase of 10 epochs, during which the learning rate is linearly increased from a starting value of $1\text{e-}7$. Following the warm-up, the model is trained for an additional 100 epochs. Unlike prior works~\citep{rlrr} that rely on grid search, we tune MoPPA's hyper-parameters, such as learning rate, drop path rate, and weight decay, based on experience. For FGVC, we applied the same dataset split implementation used in ~\citep{vpt} for a fair comparison. To align trainable parameters, for classification tasks, we additionally insert learnable global scaling \& shifting operations proposed in ~\citep{rlrr} in Multi-Layer Perceptron (MLP), patch embedding, and attention layers (only for value and output linear layers in attention operations). For detection tasks, we additionally insert a $3\times3$ convolution layer before MLP.

\section{VTAB-1K Per-Task Reults}
\label{sec:vtab_per_task}

Table~\ref{table:vtab_per_task} presents the per-task results of MoPPA on VTAB-1K, alongside baseline methods of which per-task results are available. 

\section{Results with Other Pre-trained Backbones}
\label{sec:vit_large}

To further validate the generalization of MoPPA, we evaluated its performance on diverse pre-trained backbones, including ViT-L and Swin-B.

\subsection{ViT-L}

With ImageNet-22K pre-trained ViT-L, PEFT results on VTAB-1K are summarized in Table~\ref{table:vtab_large}. One can see that MoPPA consistently achieves leading performance, validating its effectiveness on larger pre-trained vision models.

\begin{table*}[h]
	\centering
	\caption{VTAB-1K PEFT comparison with ImageNet-22K pre-trained ViT-L. Results of baseline models are obtained from ~\citep{rlrr}.}
	\resizebox{\linewidth}{!}{
		\begin{tabular}{c|ccccccc|c|cccc|c|cccccccc|c|cc}
			\toprule
			\multirow{2}{*}[-27pt]{\textbf{Methods}} & \multicolumn{8}{c|}{\textbf{Natural}}                         & \multicolumn{5}{c|}{\textbf{Specialized}} & \multicolumn{9}{c|}{\textbf{Structed}}                                &       &  \\
			 & \rotatebox{90}{\textbf{CIFAR-100}} & \rotatebox{90}{\textbf{Caltech101}} & \rotatebox{90}{\textbf{DTD}} & \rotatebox{90}{\textbf{Flowers102}} & \rotatebox{90}{\textbf{Pets}} & \rotatebox{90}{\textbf{SVHN}} & \multicolumn{1}{c}{\rotatebox{90}{\textbf{Sun397}}} & \rotatebox{90}{\textbf{Average}} & \rotatebox{90}{\textbf{Camelyon}} & \rotatebox{90}{\textbf{EuroSAT}} & \rotatebox{90}{\textbf{Resisc45}} & \multicolumn{1}{c}{\rotatebox{90}{\textbf{Retinopathy}}} & \rotatebox{90}{\textbf{Average}} & \rotatebox{90}{\textbf{Clevr-Count}} & \rotatebox{90}{\textbf{Clevr-Dist}} & \rotatebox{90}{\textbf{DMLab}} & \rotatebox{90}{\textbf{KITTI-Dist}} & \rotatebox{90}{\textbf{dSpr-Loc}} & \rotatebox{90}{\textbf{dSpr-Ori}} & \rotatebox{90}{\textbf{sNORB-Azim}} & \multicolumn{1}{c}{\rotatebox{90}{\textbf{sNORB-Ele}}} & \rotatebox{90}{\textbf{Average}} & \rotatebox{90}{\textbf{Average Total}} & \rotatebox{90}{\textbf{Params.(M)}} \\
             \midrule
		  Full fine-tuning & 68.6 & 84.3 & 58.6 & 96.3 & 86.5 & 87.5 & 41.4 & 74.7 & 82.6 & 95.9 & 82.4 & 74.2 & 83.8 & 55.4 & 55.0 & 42.2 & 74.2 & 56.8 & 43.0 & 28.5 & 29.7 & 48.1 & 65.4 & 303.4 \\
		  VPT-Deep~\cite{vpt} & 84.1 & 88.9 & 70.8 & 98.8 & 90.0 & 89.0 & 55.9 & 82.5 & 82.5 & \textbf{96.6}  & 82.6 & 73.9 & 83.9 & 63.7 & 60.7 & 46.1 & 75.7 & 83.7 & 47.4 & 18.9 & 36.9 & 54.1 & 70.8 & 0.49 \\
		  LoRA~\cite{lora}  & 75.8 & 89.8 & 73.6 & 99.1 & 90.8 & 83.2 & 57.5 & 81.4 & 86.0 & 95.0 & 83.4 & 75.5 & 85.0 & 78.1 & 60.5 & 46.7 & 81.6 & 76.7 & 51.3 & 28.0 & 35.4 & 57.3 & 72.0 & 0.74 \\
            SSF~\cite{ssf}   & 73.5 & 91.3 & 70.0 & 99.3 & 91.3 & 90.6 & 57.5 & 81.9 & 85.9 & 94.9 & 85.5 & 74.4 & 85.2 & 80.6 & 60.0 & 53.3 & 80.0 & 77.6 & 54.0 & 31.8 & 35.0 & 59.0 & 73.0 & 0.60 \\
            RLRR~\citep{rlrr} & 79.3 & 92.0 & 74.6 & 99.5 & 92.1 & 89.6 & \textbf{60.1} & 83.9 & 87.3 & 95.3 & 87.3 & 75.7 & 86.4 & \textbf{82.7} & 62.1 & 54.6 & 80.6 & 87.1 & 54.7 & 31.3 & 41.9 & 61.9 & 75.2 & 0.82 \\
            \rblue
		  \textbf{MoPPA (Ours)} & \textbf{81.6} & \textbf{95.5} & \textbf{78.3} & \textbf{99.6} & \textbf{92.5} & \textbf{92.2} & 58.9 & \textbf{85.5} & \textbf{88.3} & 96.1 & \textbf{89.1} & \textbf{76.4} & \textbf{87.5} & 80.6 & \textbf{63.7} & \textbf{54.7} & \textbf{83.6} & \textbf{88.0} & \textbf{56.2} & \textbf{33.0} & \textbf{44.3} & \textbf{63.0} & \textbf{76.5} & 0.65 \\
            \bottomrule
		\end{tabular}
	}
	\label{table:vtab_large}
\end{table*}

\begin{table*}[h]
	\centering
	\caption{VTAB-1K PEFT comparison with ImageNet-22K pre-trained Swin-B. Results of baseline models are obtained from ~\citep{rlrr}.}
	\resizebox{\linewidth}{!}{
		\begin{tabular}{c|ccccccc|c|cccc|c|cccccccc|c|cc}
			\toprule
			\multirow{2}{*}[-27pt]{\textbf{Methods}} & \multicolumn{8}{c|}{\textbf{Natural}}                         & \multicolumn{5}{c|}{\textbf{Specialized}} & \multicolumn{9}{c|}{\textbf{Structed}}                                &       &  \\
			 & \rotatebox{90}{\textbf{CIFAR-100}} & \rotatebox{90}{\textbf{Caltech101}} & \rotatebox{90}{\textbf{DTD}} & \rotatebox{90}{\textbf{Flowers102}} & \rotatebox{90}{\textbf{Pets}} & \rotatebox{90}{\textbf{SVHN}} & \multicolumn{1}{c}{\rotatebox{90}{\textbf{Sun397}}} & \rotatebox{90}{\textbf{Average}} & \rotatebox{90}{\textbf{Camelyon}} & \rotatebox{90}{\textbf{EuroSAT}} & \rotatebox{90}{\textbf{Resisc45}} & \multicolumn{1}{c}{\rotatebox{90}{\textbf{Retinopathy}}} & \rotatebox{90}{\textbf{Average}} & \rotatebox{90}{\textbf{Clevr-Count}} & \rotatebox{90}{\textbf{Clevr-Dist}} & \rotatebox{90}{\textbf{DMLab}} & \rotatebox{90}{\textbf{KITTI-Dist}} & \rotatebox{90}{\textbf{dSpr-Loc}} & \rotatebox{90}{\textbf{dSpr-Ori}} & \rotatebox{90}{\textbf{sNORB-Azim}} & \multicolumn{1}{c}{\rotatebox{90}{\textbf{sNORB-Ele}}} & \rotatebox{90}{\textbf{Average}} & \rotatebox{90}{\textbf{Average Total}} & \rotatebox{90}{\textbf{Params.(M)}} \\
             \midrule
		  Full fine-tuning & 72.2 & 88.0 & 71.4 & 98.3 & 89.5 & 89.4 & 45.1 & 79.1 & 86.6 & \textbf{96.9} & 87.7 & 73.6 & 86.2 & 75.7 & 59.8 & 54.6 & 78.6 & 79.4 & 53.6 & \textbf{34.6} & \textbf{40.9} & 59.7 & 72.4 & 86.9 \\
		  VPT-Deep~\cite{vpt} & \textbf{79.6} & 90.8 & \textbf{78.0} & 99.5 & 91.4 & 46.5 & 51.7 & 76.8 & 84.9 & 96.2 & 85.0 & 72.0 & 84.5 & 67.6 & 59.4 & 50.1 & 74.1 & 74.4 & 50.6 & 25.7 & 25.7 & 53.4 & 67.7 & 0.22 \\
            RLRR~\citep{rlrr} & 66.1 & 90.6 & 75.5 & 99.3 & \textbf{92.1} & \textbf{90.9} & \textbf{54.7} & 81.3 & 87.1 & 95.9 & 87.1 & 76.5 & 86.7 & 66.0 & 57.8 & 55.3 & \textbf{84.1} & \textbf{91.1} & \textbf{55.2} & 28.6 & 34.0 & 59.0 & 73.0 & 0.41 \\
            \rblue
		  \textbf{MoPPA (Ours)} & 70.7 & \textbf{93.9} & 76.2 & \textbf{99.7} & 92.0 & 90.0 & \textbf{54.7} & \textbf{82.5} & \textbf{87.3} & 96.1 & \textbf{88.4} & \textbf{76.9} & \textbf{87.2} & \textbf{78.1} & \textbf{59.9} & \textbf{56.1} & 84.0 & 87.6 & 54.9 & 32.2 & 38.4 & \textbf{61.4} & \textbf{74.6} & 0.39 \\
            \bottomrule
		\end{tabular}
	}
	\label{table:vtab_swinb}
\end{table*}

\subsection{Swin-B}

Additionally, we tested its performance with ImageNet-22K pre-trained Swin-B, and results are provided in Table~\ref{table:vtab_swinb}. MoPPA achieves outperforming results compared with other baseline methods, validating its generalization across diverse vision representation backbones.

\section{Detailed Analysis Implementation}
\label{sec:analysis_implementation}

We conduct experiments to evaluate the adaptation capacity, with a randomly generated input tensor ($14\times14\times768$) and a randomly generated corresponding Ground Truth (GT) tensor ($14\times14\times768$). All values are sampled from a uniform distribution $\text{U(0,1)}$. We then fine-tune a pre-trained ViT model for 20000 iterations by using the AdamW optimizer~\citep{adamw} with a learning rate $0.002$, equipped with LoRA (taking $rank=6$ to align trainable parameters) / MoPPA, to predict the GT with the input tensor, supervised by Mean Squared Error (MSE). The results of 5 trials are reported in Table~\ref{table:detail_adaptation}.

\begin{table}[h]
\centering
\caption{MSE of 5 trials for LoRA and MoPPA.}
\begin{tabular}{l|cc}
     \toprule
     Trial & LoRA & \textbf{MoPPA (Ours)} \\
     \midrule
     1 & 0.0648 & \textbf{0.0507} \\ 
     2 & 0.0645 & \textbf{0.0513} \\
     3 & 0.0636 & \textbf{0.0510} \\
     4 & 0.0656 & \textbf{0.0500} \\
     5 & 0.0649 & \textbf{0.0486} \\
     \bottomrule
\end{tabular}
\label{table:detail_adaptation}
\end{table}

\section{Visualization of Coefficients in Physical Equations}
\label{sec:vis_weights}

\begin{figure}[h]
    \centering
    \includegraphics[width=0.995\linewidth]{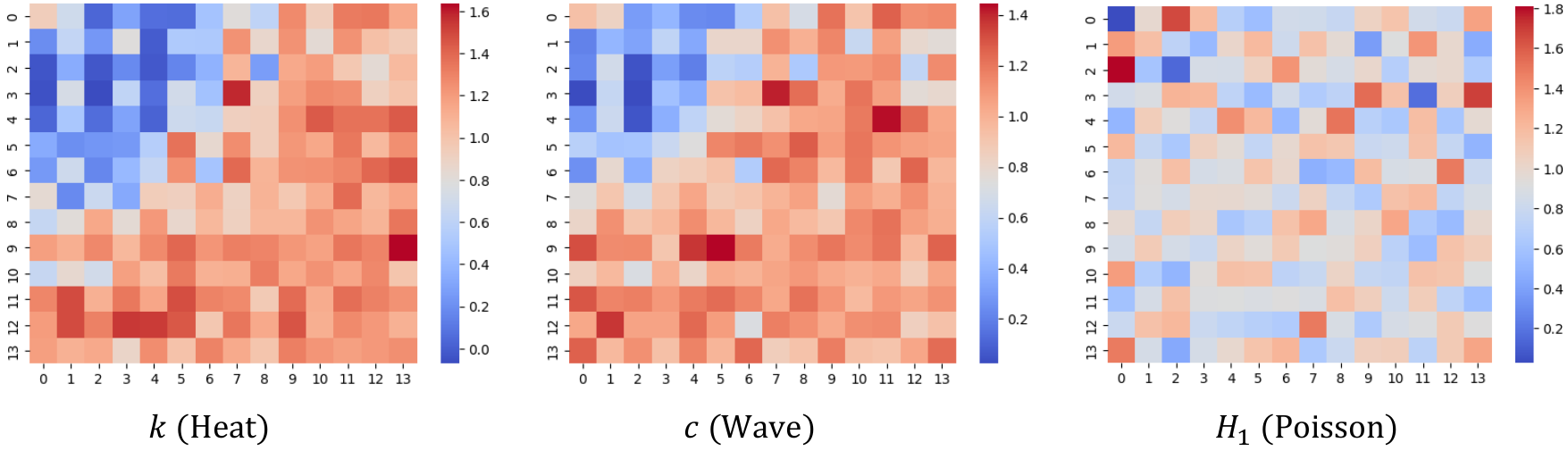}
    \caption{Visualization of $k/c/H_1$ proposed in Sec.~\ref{sec:moppa} in $\text{Heat()/Wave()/Poisson()}$ implementations with DCT domain coordinates.}
    \label{fig:vis_weights}
\end{figure}

The visualization of \( k/c/H_1 \) proposed in Sec.~\ref{sec:moppa} for the implementations of \(\text{Heat()} / \text{Wave()} / \text{Poisson()}\) is presented in Fig.~\ref{fig:vis_weights}. From the figure, we observe that \( k/c \) in \(\text{Heat()}\) and \(\text{Wave()}\) exhibit lower or higher values corresponding to lower or higher frequency regions, respectively. According to the underlying equations, higher \( k/c \) values represent lower frequency filtering coefficients. This trend, as visualized, supports the interpretation that \(\text{Heat()}\) and \(\text{Wave()}\) in the proposed MoPPA act as adaptive low-frequency enhancement filters. In contrast, \( H_1 \) in \(\text{Poisson()}\) displays a more random structure with a banded pattern. Through the weighted combinations determined by the router in each block, MoPPA effectively facilitates parameter-efficient fine-tuning of pre-trained models by leveraging diverse physical priors. This adaptive mechanism enables tailored feature transformations to align with various tasks.

\end{document}